\documentclass[letterpaper, 10 pt, conference]{ieeeconf}  

\IEEEoverridecommandlockouts                              

\usepackage{graphics} 
\usepackage{epsfig} 
\usepackage{mathptmx} 
\usepackage{times} 
\usepackage{amsmath} 
\usepackage{amssymb}
\usepackage{pifont}
\newcommand{\cmark}{\ding{51}}%
\newcommand{\xmark}{\ding{55}}%
\usepackage{amssymb}  
\usepackage{bm}
\usepackage{algorithm}
\usepackage{algpseudocode}
\usepackage{tabulary}
\usepackage{caption}
\usepackage{subcaption}
\usepackage{multirow}
\usepackage{euscript}
\usepackage{tikz}
\usepackage{cite}
\usepackage{color, xcolor}

\usepackage[english]{babel}
\usepackage[T1]{fontenc}
\usepackage[utf8]{inputenc}
\usepackage{booktabs} 
\usepackage{colortbl}

\usepackage{hyphenat}
\usepackage[hidelinks]{hyperref} 
\usepackage{hyperref}

\usepackage[nopostdot,nogroupskip,nonumberlist]{glossaries}
\newglossaryentry{ad}{name=AD, description={Anomaly Detection},first={\textit{Anomaly Detection} (AD)}}
\newglossaryentry{asp}{name=ASP, description={Assembly Sequence Planning},first={\textit{Assembly Sequence Planning} (ASP)}}
\newglossaryentry{auc}{name=AUC, description={Area Under Curve},first={\textit{Area Under Curve} (AUC)}}
\newglossaryentry{aco}{name=ACO, description={Ant Colony Optimization},first={\textit{Ant Colony Optimization} (ACO)}}

\newglossaryentry{cad}{name=CAD, description={Computer-Aided Design},first={\textit{Computer-Aided Design} (CAD)}}

\newglossaryentry{dof}{name=DoF, description={Degrees of Freedom},first={\textit{Degrees of Freedom} (DoF)}}
\newglossaryentry{dfs}{name=DFS, description={Degrees of Freedom},first={\textit{Depth First Search} (DFS)}}

\newglossaryentry{ea}{name=EA, description={Evolutionary Algorithms},first={\textit{Evolutionary Algorithms} (EA)}}

\newglossaryentry{gnn}{name=GNN, description={Graph Neural Network},first={\textit{Graph Neural Networks} (GNNs)}}
\newglossaryentry{gan}{name=GAN, description={Generative Adversarial Network},first={\textit{Generative Adversarial Network} (GAN)}}
\newglossaryentry{gcn}{name=GCN, description={Graph Convolution Network},first={\textit{Generative Adversarial Network} (GCN)}}
\newglossaryentry{gat}{name=GAT, description={Graph Attention Network},first={\textit{Graph Attention Network} (GAT)}}
\newglossaryentry{ga}{name=GA, description={Genetic Algorithms},first={\textit{Genetic Algorithms} (GA)}}
\newglossaryentry{gin}{name=GIN, description={Graph Isomorphism Network},first={\textit{Graph Isomorphism Network} (GIN)}}

\newglossaryentry{ir}{name=IR, description={Information Retrieval},first={\textit{Information Retrieval} (IR)}}
\newglossaryentry{il}{name=IL, description={Imitation Learning},first={\textit{Imitation Learning} (IL)}}

\newglossaryentry{kde}{name=KDE, description={Kernel Density Estimation},first={\textit{Kernel Density Estimation} (KDE)}}

\newglossaryentry{mdp}{name=MDP, description={Markov Decision Process},first={\textit{Markov Decision Process} (MDP)}}

\newglossaryentry{nn}{name=NN, description={Neural Network},first={\textit{Neural Network} (NN)}}

\newglossaryentry{pk}{name=P@k, description={Precision@k},first={\textit{Precision@k} (P@k)}}

\newglossaryentry{rasp}{name=RASP, description={Robotic Assembly Sequence Planning},first={\textit{Robotic Assembly Sequence Planning} (RASP)}}
\newglossaryentry{rl}{name=RL, description={Reinforcement Learning},first={\textit{Reinforcement Learning} (RL)}}
\newglossaryentry{roc-auc}{name=ROC AUC, description={Area Under the Receiver Operating Characteristic Curve},first={\textit{Area Under the Receiver Operating Characteristic Curve} (ROC AUC)}}
\newglossaryentry{roc}{name=ROC, description={Receiver Operating Characteristic},first={\textit{Receiver Operating Characteristic} (ROC)}}
\newglossaryentry{rk}{name=R@k, description={Recall@k},first={\textit{Recall@k} (R@k)}}
\newglossaryentry{rbf}{name=RBF, description={Radial Basis Function},first={\textit{Radial Basis Function} (RBF)}}

\newglossaryentry{svm}{name=SVM, description={Reinforcement Learning},first={\textit{Reinforcement Learning} (SVM)}}

\newglossaryentry{tamp}{name=TAMP, description={Support Vector Machines},first={\textit{Support Vector Machines} (TAMP)}}

\newcommand{\etal}{\textit{et al.}~}
\newcommand{\eg}{e.g.,~}

\title{\LARGE \bf
		Efficient and Feasible Robotic
		Assembly Sequence Planning via
		\\  Graph Representation Learning
}

\author{Matan~Atad$^{\star 2}$, Jianxiang~Feng$^{\star 1,2}$, Ismael~Rodríguez$^{1,2}$, Maximilian~Durner$^{1,2}$ and Rudolph~Triebel$^{1,2}$
\thanks{$^{\star}$ Equal Contribution.}%
\thanks{$^{1}$ Institute of Robotics and Mechatronics, German Aerospace Center (DLR), 82110 Wessling, Germany. {\tt\small <first>.<second>@dlr.de}}%
\thanks{$^{2}$ Department of Informatics, Technical University of Munich, 85748 Garching, Germany. {\tt\small <first>.<second>@tum.de}}
}

\makeatletter
\def\endthebibliography{%
	\def\@noitemerr{\@latex@warning{Empty `thebibliography' environment}}%
	\endlist
}

\begin{document}

\maketitle

\begin{abstract}
Automatic Robotic Assembly Sequence Planning (RASP) can significantly improve productivity and resilience in modern manufacturing along with the growing need for greater product customization.
One of the main challenges in realizing such automation resides in efficiently finding solutions from a growing number of potential sequences for increasingly complex assemblies.
Besides, costly feasibility checks are always required for the robotic system.
To address this, we propose a holistic graphical approach including a graph representation called Assembly Graph for product assemblies and a policy architecture, Graph Assembly Processing Network, dubbed GRACE for assembly sequence generation.
With GRACE, we are able to extract meaningful information from the graph input and predict assembly sequences in a step-by-step manner.
In experiments, we show that our approach can predict feasible assembly sequences across product variants of aluminum profiles based on data collected in simulation of a dual-armed robotic system. 
We further demonstrate that our method is capable of detecting infeasible assemblies, substantially alleviating the undesirable impacts from false predictions, and hence facilitating real-world deployment soon.
Code and training data are available at \href{https://github.com/DLR-RM/GRACE}{https://github.com/DLR-RM/GRACE}.
\end{abstract}

\section{Introduction}
Aiming for high flexibility, manufacturers around the globe are introducing automation for \gls{rasp} at a greater pace to respond to rapid changes in market needs for customization of novel product variants~\cite{Shih2020}. 
These changes cause often modifications in assembly lines, requiring time-consuming and resource-intensive re-planning,
because of the NP-hard combinatorial characteristic~\cite{Rashid2011} of \gls{asp}, where the number of possible solutions grows with the factorial of the amount of parts involved.
Also, to check whether a certain assembly sequence can actually be executed on a specific robotic system is computationally expensive.
For example in~\cite{suarez2018can}, 11 minutes were required for the assembly motion planning of an IKEA chair.
This is more time than the actual execution of the plan, not to mention the case of product variants whose assembly sequence space itself must be explored, easily leading to a search of hours or days instead of minutes.
\begin{figure}[t]
	\centering
	\includegraphics[width=1.\linewidth, height=8.6cm]{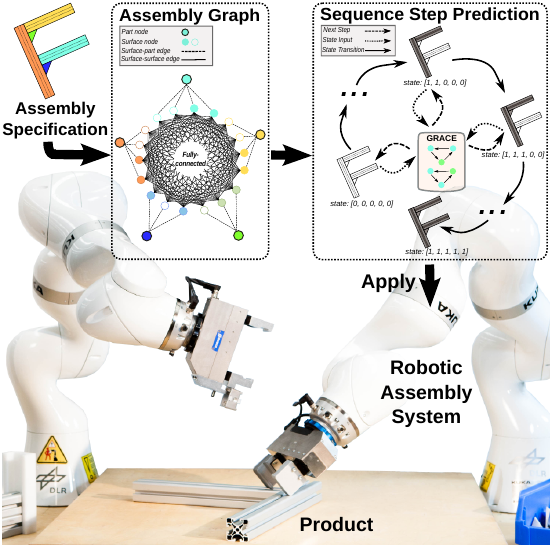}
	\caption{\textbf{Workflow of our proposed graphical ASP method} on a dual-armed robotic system (simulated in our setting): an aluminum assembly specification is first represented as an Assembly Graph and then fed into our policy network GRACE, designed to flexibly and efficiently generate assembly sequences in a \textit{step-by-step} manner that can be executed by the robot. Best viewed in color.}
	\label{fig:teaser}
\end{figure}

Several existing works attempt to improve the tedious \gls{asp} process by predicting feasible assembly sequences~\cite{zhao2019aspw,watanabe2020search} or inferring the underlying rules guiding their creation~\cite{rodriguez2019iteratively, rodriguez2020pattern}.
Although those works already facilitate the assembly planning, they still lack desirable attributes such as generalization across varying product types and sizes as well as run-time efficiency.
In this work, we address these problems with a graphical learning-based approach, that is able to automatically generate sequences for \textit{unknown} assembly variants in an \emph{efficient} way.

In a nutshell of our main idea, inspired by~\cite{lin2022efficient}, we formulate \gls{rasp} as a sequential decision-making problem with a \gls{mdp}, in order to break the restriction of combinatorial complexity wrt. the number of parts and thus, boost generalization performance.
Hence the sequence is generated step-by-step based on the current assembly state.
Meanwhile we exploit the idea of distilling previous knowledge acquired for assembling products to predict the next feasible actions with a designed policy architecture. This architecture is optimized to imitate the demonstration sequences collected in simulation which are interpreted as expert demonstrations. 

Specifically, to put the aforementioned ideas into practice, we propose to use a graphical representation to faithfully describe the spatial structure of assemblies.
Our so-called Assembly Graph is adapted from and more fine-grained than the one in~\cite{rodriguez2020pattern} by representing the assembly as a heterogeneous graph whose edges denote geometrical relations between the assembly part surfaces.
Based on this, we further develop a policy architecture based on \gls{gnn}, called \textbf{GR}aph \textbf{A}ssembly pro\textbf{C}essing n\textbf{E}tworks, for short GRACE, to extract useful information from the Assembly Graph and predict actions determining which parts should be assembled next.
Apart from this, false predicted sequences and infeasible assemblies pose a severe problem for efficiency of learning-based assembly robots, \eg an incorrect sequence might require the robot to perform time-consuming re-planning.
Therefore, it would be beneficial to detect these beforehand, \eg being introspective against false predictions\cite{feng2022introspective}, hence we further develop and analyze various schemes to enhance the performance of feasibility prediction. 

It is worthwhile to note that there are several advantages for the proposed graphical representation and the policy architecture, GRACE:
(1) Invariance to number of parts: contrary to previous works such as~\cite{rodriguez2020pattern} restricted by a fixed number of parts, ours is free from this limitation, as 
GRACE is capable of handling varying number of input graph nodes.
(2) Memory efficient learning: GRACE 
employs shared weights across all nodes in the graph, further alleviating the burden of the aforementioned complexity.
(3) Generalization: GRACE trained on assemblies of one size is able to generalize to those of smaller sizes (see results in Tab.~\ref{exp:tab_knowledge_transfer_inter_sized_one2many}). 
(4) Multiple solutions: GRACE predicts several feasible sequences (in contrast to~\cite{lin2022efficient}), allowing greater flexibility and resilience during execution.

We validate the proposed method with comprehensive experiments based on a dataset of assemblies made of different numbers of aluminum parts created in simulation of a dual-armed robotic system.
This setting can be mapped to various tasks in the industry~\cite{rodriguez2020pattern} as it allows for construction of numerous product variations.
Moreover, as shown in Fig.~\ref{fig:asp_exp_aluminum_assemblies}, it requires a deeper understanding of several complex relations (\eg distances between parts, physical part characteristics). 
The results show that our approach is able to efficiently predict feasible assembly sequences across product variants (with few millisecond to predict the next step (\ref{implementation})).

To summarize, our contribution is three-fold:
\begin{itemize}
	\item We introduce Assembly Graph, a heterogeneous graphical representation for the \gls{rasp} task, which is a more fine-grained and flexible representation than our previous one in~\cite{rodriguez2020pattern} by including part surfaces and parts in the same graph. 
	\item We develop a policy architecture GRACE to process the Assembly Graph and predict feasible assembly sequences in a step-by-step manner as well as the feasibility for a given assembly specification.
	\item We conduct comprehensive experiments in simulation to validate the proposed approach including failure analysis and ablation studies on design choices.
\end{itemize}

\begin{table*}[t]
	\centering
 	\caption{Comparison between our proposed method and other relevant works.}
	\resizebox{\textwidth}{!}{
	\begin{tabular}{l c c c c }
	\toprule
	 & \shortstack{Efficient generalization \\across assembly sizes?}& \shortstack{Robotic constraints \\ involved?}  & \shortstack{Direct sequences \\generation?} & \shortstack{Fine-grained \\graph representation?} \\ 	
	 \midrule
	ASPW-DRL \cite{zhao2019aspw}	& \xmark & \xmark & \cmark & \xmark \\ 
	LEGO-GRAPH \cite{ma2022planning}	& \cmark & \xmark & \cmark & \xmark \\		
	KT-RASP \cite{rodriguez2020pattern}	& \xmark & \cmark & \xmark & \xmark \\ 	
	Our proposed method	& \cmark & \cmark & \cmark & \cmark \\ 	
	\bottomrule
	\end{tabular}
	}
    \label{tab:comparison}
\end{table*}
\section{Related Work}

\paragraph{Assembly Sequence Planning} 
A popular assembly graph representation for \gls{asp} is the AND/OR Graph~\cite{54734}, a formalism to encode the space of feasible assembly sequences, which can be created with the Disassembly For Assembly strategy~\cite{1087132, 1217194, 7559140, 7294142}. 
However, these approaches are restricted on time to find a solution efficiently due to the feasibility checks.
While graph search methods are impractical for larger assemblies because of the combinatorial explosion problem, heuristic intelligent search methods provide another alternative.
They reject infeasible sequences and search for feasible ones close to the optimal based on manually designed termination criteria\cite{9762444, Iwankowicz2016}, learned~\cite{chen2008three,sinanouglu2005assembly} or hand-crafted~\cite{Rashid2017} energy functions.
More recently, Zhao \etal\cite{zhao2019aspw} and Watanabe \etal\cite{watanabe2020search} applied deep \gls{rl} for \gls{asp}.
Different to us, they do not have a graph representation to take into account relations between parts.
Targeting at \gls{rasp}, Rodriguez \etal\cite{rodriguez2019iteratively, rodriguez2020pattern} suggested inferring assembly rules (\eg a specific part should be assembled before another), which can be transferred from previous identified sub-assemblies to those of larger sizes to prune the search space, thus reducing planning time.
Their approach only produces rules, from which the final assembly sequences need to be derived additionally.
It also requires further re-training when adapting to other product variants.
Enlightened by them, we refine their graph representation to a more fine-grained level and adapt their idea with a learning-based approach, aiming to mitigate these issues. 
Similar to us, Ma \etal\cite{ma2022planning} used \gls{gnn} for \gls{asp} of LEGO structures. 
However, they differ from us in two aspects, first they do not consider assembly robots in the loop and second they model assemblies only with a coarser graph representation whose edges only consider connections among parts instead of part surfaces.
To clearly show different characteristics among relevant works, we provide a concise comparison in Tab. \ref{tab:comparison}.

\paragraph{Graph Representation Learning in Task Planning} 
In this setting, graphs commonly incorporate nodes for manipulated objects~\cite{nguyen2020self,bapst2019structured,zhu2021hierarchical}, their target positions~\cite{lin2022efficient,funk2022learn2assemble} and the robot gripper~\cite{ye2020object}. 
Edges can represent high-level relations between objects~\cite{nguyen2020self,zhu2021hierarchical}.
With the graph representation, Zhu \etal\cite{zhu2021hierarchical} and Ye \etal\cite{ye2020object} generated feasible candidate paths by sampling, and trained a network that predicts a sequence of feasible actions in backward and forward search, respectively. 
Nguyen \etal\cite{nguyen2020self} performed sampling to find action sequences that transform the source to target graph and then used optimization to eliminate invalid sequences subject to the environment constraints. 
Besides, some researchers resorted to \gls{rl}~methods such as~\cite{bapst2019structured,funk2022learn2assemble}, and \cite{li2020towards}, who used \gls{gnn}s for task planning.
Recently, Lin \etal\cite{lin2022efficient} utilized \gls{il} to train two \gls{gnn}s, one for selecting objects in the scene and another picking a suitable goal state from a set of possible goal positions for long-horizon manipulation tasks.
Inspired by them, we train our \gls{gnn}s for \gls{rasp} task by leveraging \gls{il} for ease and efficiency in training.

\section{Background}
In this section, we briefly recap the concept of \glspl{gnn} and heterogeneous graphs, which are the base for our method.

\paragraph{Graph Neural Networks} A \gls{gnn} operates on an undirected graph $\mathcal{G}=(\mathcal{V},\mathcal{E})$ with nodes $\mathcal{V}$ and edges $\mathcal{E}$, where every node $v \in \mathcal{V}$ is assigned with a feature vector $\phi(v)$. 
It updates node features by exchanging information between neighboring nodes. 
This is done with multiple Message Passing layers~\cite{gilmer2017neural}. 
For each layer $l$, let $\mathbf{h}_i^0 = \phi(v_i)$ be the input features of node $v_i$ and $\mathcal{N}_i$ its set of neighboring nodes. Then we can define a three-step process to update these features:
\begin{enumerate}
	\item \emph{Gather} feature from neighboring nodes: ${{\{\mathbf{h}_j^{l-1}\}}_{j \in \mathcal{N}_i}}$.
	\item \emph{Aggregate} messages from the neighboring nodes: ${\mathbf{m}_i^l = g_{\omega}({\{\mathbf{h}_j^{l-1}\}}_{j \in \mathcal{N}_i})}$.
	\item \emph{Update} features of node $v_i$: ${\mathbf{h}_i^l = f_{\phi}^l(\mathbf{h}_i^{l-1}, \mathbf{m}_i^l)}$.
\end{enumerate}
The function $g_{\omega}$ can be either constant (e.g.\ sum) or learned during training.
The term $f_{\phi}$ is a \gls{nn} parameterized by $\phi$.
Both, $f_{\phi}$ and $g_{\omega}$, are shared across all nodes in the graph, making \gls{gnn}s efficient and independent of the number of nodes in the graph. 

In our proposed method we apply a Graph Attention Network (GAT)~\cite{velivckovic2017graph}, a popular variant of \gls{gnn}s, that defines $g_{\omega}$ as attention:
\begin{align}
\label{eq:2}
\mathbf{m}_i^l &= \sum_{j \in \mathcal{N}_i} \left( \alpha_{i,j} \cdot \mathbf{h}_j^{l-1} \right), \\
\mathbf{h}_i^l &=  \mathbf{W}_1 \cdot \alpha_{i,i}  \mathbf{h}_i^{l-1} + \mathbf{W}_1 \cdot \mathbf{m}_i^l, \\
\label{eq:3}
\alpha_{i,j} &= \frac{\exp \left(\mathbf{a} \cdot \sigma \left( \mathbf{W}_2 [\mathbf{h}_i^{l-1} \mathbin\Vert \mathbf{h}_j^{l-1} \mathbin\Vert e_{i,j}] \right) \right)}{ \sum_{k \in \mathcal{N}_i \cup \{ i \}} \exp \left(\mathbf{a} \cdot \sigma \left( \mathbf{W}_2 [\mathbf{h}_i^{l-1} \mathbin\Vert \mathbf{h}_k^{l-1} \mathbin\Vert e_{i,k}] \right) \right)},
\end{align}
where $\mathbf{W}_1$, $\mathbf{W}_2$, and $\mathbf{a}$ are learned, $\sigma$ is a Leaky ReLU activation function, and $[a \mathbin\Vert b]$ is a concatenation operator between $a$ and $b$.

\paragraph{Heterogeneous Graph} $\mathcal{G}=(\mathcal{V},\mathcal{E})$ generalizes graphs to multiple types of nodes and edges~\cite{sun2013mining}. 
Each node $v \in \mathcal{V}$ belongs to one particular node type $\psi_n(v)$ and analogously each edge $e \in \mathcal{E}$ to an edge type $\psi_e(e)$.
In~\cite{wang2019heterogeneous}, the authors extend \glspl{gat} to a heterogeneous graph setting.
This is accomplished by obtaining for each node a different updated feature vector per group of specific neighboring source node and edge types, and aggregating the features to obtain a single result, for instance using a sum. 
This formulation is essential, as every type of neighboring node may have a different feature dimension.

\section{Method}
In this section, \gls{rasp} is formulated as a sequential decision-making problem with a \gls{mdp} and then we present our graph representation to depict assemblies. 
Based on this, we elaborate the proposed network GRACE, and demonstrate the assembly sequence generation.

\subsection{Problem Formulation}\label{sec:problem_formulation}
We describe the sequence prediction task for an assembly with $N$ parts as a \gls{mdp}~\cite{bellman1957markovian} with a discrete state space $\mathcal{S}$ and a high-level discrete action space $\mathcal{A}$.

Starting from state $\textbf{s}_t$ at time step $t$, executing action $a_t$ produces a reward $r_t$ and switches to state $\textbf{s}_{t+1} \sim p(\textbf{s}_{t+1}|\textbf{s}_t, a_t)$ with a transition function $\textit{p}$.
State $\textbf{s}_t \in {\{0, 1\}}^N$ is a binary vector indicating which parts are already placed in their target position by $1$ (i.e.\ assembled) otherwise by $0$. 
Action $a_t \in \{1, \ldots ,N\}$ represents the next part placement among the unplaced ones.
For \emph{feasible} assemblies, there are multiple different sequences leading to the final state, in which all $N$ parts are placed correctly.
For \emph{infeasible} assemblies, no sequence exists, due to constraints of different aspects spanning from part geometries to kinematic and dynamics regarding the robotic system.
Our objective is to learn a policy network $\pi_{\theta}(\textbf{s}_{t}) = a_t$ parameterized by $\theta$, which is optimized to imitate the assembly demonstrations 
${\tau_i=\{\textbf{s}_{i,1}, a^{exp}_{i,1}, \ldots, \textbf{s}_{i,T}, a^{exp}_{i,T}\}}$ in a dataset of $M$ sequences ${\mathcal{D} = {\{\tau_i\}}^{M}_{i=1}}$ and generalize across variants of different types and sizes at test time. 
In practice, our network predicts a set of multiple possible actions e.g. $K_t=\{a_{t,k}\}_{k=1}^{|K_t|}$ based on a tunable threshold to control the prediction quality. 

\subsection{Assembly Graphs}\label{method:aseembly_graphs}
We represent the overall structure of an assembly with a heterogeneous graph. 
To make this representation agnostic to the rotation and mirroring of the assembly structure, we employ only relative distances instead of absolute positions for the features of edges between surfaces.
More formally, given an assembly $A$ (Fig.~\ref{fig:GRACE_fig}) at state $\textbf{s}_t$ it is modeled as a graph ${\mathcal{G}_t=(\mathcal{V},\mathcal{E})}$ containing two types of nodes: part nodes $\mathcal{V}^p$ and surface nodes $\mathcal{V}^s$, and two types of edges: $\mathcal{E}^{s \text{-to-} s}$, connecting all surface nodes, and $\mathcal{E}^{s \text{-to-} p}$, connecting each surface node to its respective part. 
We detail each component as follows:
\subsubsection{Part Nodes} 
Responsible for encoding the current state of the assembly. A part node $\textit{v}^p_i \in \mathcal{V}^p$ is associated with a feature vector $\phi(\textit{v}_i^p)= [\textit{assembled-flag} \in \{ 0, 1 \},\ \textit{part-type} \in \mathbb{N},\ \textit{part-id} \in \mathbb{R}^d]$.
There are three atomic part types: \emph{long profile}, \emph{short profile} and \emph{angle bracket}.

\subsubsection{Surface Nodes} 
Different to the one in \cite{rodriguez2020pattern}, we associate each surface node $\textit{v}^s_i \in \mathcal{V}^s$ with the features $\phi(\textit{v}_i^s)= [\textit{surface-type} \in \mathbb{N},\ \textit{surface-id} \in \mathbb{R}^d]$. 
There are two surface types (\emph{long} and \emph{short}) for profiles and one (\emph{lateral}) for brackets.
Both the \textit{part-id} and \textit{surface-id} fields are encoded with a $d$-dimensional Sinusoidal Positional Encoding~\cite{vaswani2017attention}. 

\subsubsection{Surface-to-Surface Edges} 
We design a fully-connected graph for all surface nodes $\mathcal{V}^s$ to capture the relation between untouched surfaces, which is more fine-grained than those in \cite{rodriguez2020pattern} with only connects between touched surfaces.
These edges are assigned with a feature $\phi(\textit{e}_i) \in \mathbb{R}$, indicating the \emph{relation} between the two surfaces: 
$\phi(\textit{e}_i)=$ \textit{relative distance} (parallel); $1$ (belong to the same part); $-1$ (orthogonal); $0$ (same-surface loop).

\subsubsection{Surface-to-Part Edges} 

These connect each surface and part node pair 
$(\textit{v}^s_i,\textit{v}^p_j) \in \mathcal{V}^s \times \mathcal{V}^p$, where surface $\textit{v}^s_i$ belongs to the part $\textit{v}^p_j$.
This type of edges is not associated with any feature vector.

\subsection{Graph Assembly Processing Networks (GRACE)}\label{GRACE}
\begin{figure}[bt]
	\centering
	\includegraphics[width=1.\linewidth, height=4.7cm]{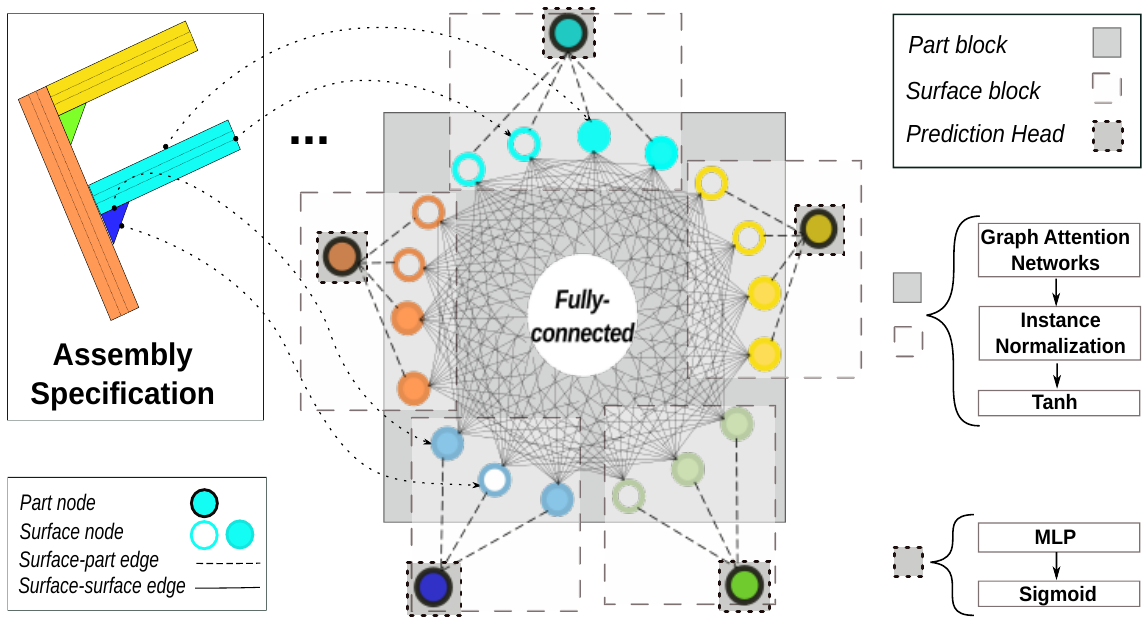}
	\caption{\textbf{Illustration of Assembly Graph and GRACE.}
		Assembly Graph consists of edges connecting parts and their surfaces and edges among all part surfaces. In GRACE, the Part Block is shared for sub-graphs of part surfaces and the attached part, while Surface Block is for the sub-graph of all part surfaces.
	To predict scores for parts to be assembled next, we apply a prediction head on each spare part.
}
	\label{fig:GRACE_fig}
\end{figure}

Based on the formulation of a \textit{step-by-step} sequential decision-making process per each part in the assembly in \ref{sec:problem_formulation}, we introduce \textbf{GR}aph \textbf{A}ssembly pro\textbf{C}essing n\textbf{E}tworks, for short GRACE, $\pi_{\theta}: \mathcal{S} \rightarrow \mathcal{A}$, where $a_i = {\{y_i | y_i \geq \lambda \}}_{i=1}^N$, to extract useful information from the Assembly Graph and predict the next action given the current state of an assembly of $N$ parts. $\lambda \geq 0$ is a threshold used to control the quality of predicted sequences.
GRACE outputs a score per part $y_i \in [0,1], i \in \{1, \ldots ,N\}$, reflecting the probability of placing the $i$-th part next.
We further articulate the main components of this network (Fig.~\ref{fig:GRACE_fig}),
describe the algorithm for predicting the entire sequence of length $N$ by traversing predicted steps and the way we infer the feasibility of a given assembly.

\subsubsection{Surface and Part Blocks}\label{model_blocks}

The architecture is made of identical blocks, which are applied sequentially to obtain updated node features. 
Each block is made of a \gls{gat}~\cite{velivckovic2017graph}, an Instance Normalization layer~\cite{ulyanov2016instance} and a Tanh function.
We choose \gls{gat} as it allows to utilize the rich semantics of edge features for updating node features in our graph representation. 
Surface Blocks are applied on surface nodes $\mathcal{V}^s$ and surface-to-surface edges $\mathcal{E}^{s \text{-to-} s}$ for updating surface node features $\phi(\textit{v}_i^s)$,  
while Part Blocks are applied on surface nodes $\mathcal{V}^s$, part nodes $\mathcal{V}^p$ and surface-to-part edges $\mathcal{E}^{s \text{-to-} p}$ to update part node features $\phi(\textit{v}_i^p)$.

\subsubsection{Prediction Head and Loss Function}

To obtain a score per part, a fully-connected layer followed by a Sigmoid function is applied on each part node.
During training, we minimize the loss between the network outputs and the ground-truth sequence steps from a dataset of assembly sequences (see ~\ref{sec:problem_formulation}) using binary cross-entropy.
To note that, we apply this loss function for each part node separately. 
Our objective function (\ref{eq:03_loss}) includes an additional regularization term (\ref{eq:03_reg}), aiming at encouraging the network not to predict already placed parts:
\begin{align} 
L_{\theta} &= \sum_{i=1}^M \sum_{j=1}^{N_i} \left( \widehat{y}_{ij} \cdot \log (y_{ij}) + (1 - \widehat{y}_{ij}) \log (1 - y_{ij})\right) + \delta L_{\text{reg}}, \label{eq:03_loss} \\
L_{\text{reg}} &= \sum_{i=1}^M \sum_{j=1}^{N_i} f_{ij} \cdot y_{ij} \label{eq:03_reg},
 \end{align}
where $M$ is the number of data examples in the dataset, $N_i$ is the number of nodes in the $i$-th graph. Abusing the notations, we denote $y_{ij}$ and $\widehat{y}_{ij}$ the output score of the model $\pi_{\theta}$ and the ground-truth step in a sequence for the $j$-th node in the $i$-th graph respectively. 
$\delta$ is a weighing coefficient and $f_{ij}$ the value of the \emph{assembled-flag} in the input features.

\subsubsection{Predicting Sequences}
As described, GRACE predicts a set of possible next steps based on the current state of an assembly.
In order to generate a complete sequence (i.e.\ of length $N$), we repeatedly apply GRACE based on the current predicted state of the Assembly Graph. 
We devise an algorithm (Algo.~\ref{alg:03_walk_tree}) to traverse the assembly state tree using \gls{dfs}: 
 
Starting with the graph in its initial state $\mathcal{G}_0$ -- for all part nodes, \emph{assembled-flag}s are set to zero, the algorithm performs the following steps recursively: 
First, it checks the exit condition of the recursion -- if all parts are already in place. 
Next, it predicts the probability for each part node $y_{i}$ and picks those larger than the threshold $\lambda$, controlling the trade-off between precision and recall. 
Each of those nodes spawns a new branch individually.
Therefore, we set the \emph{assembled-flag} and call the recursion on the altered graph to retrieve possible sequences starting with the chosen node. 
Finally, we add the chosen nodes to the head of each returned sequence and return.

\begin{algorithm}[bt]
	\caption{Assembly State Tree Traversal}\label{alg:03_walk_tree}
	\begin{algorithmic}
		\Function{Traverse-Tree}{Model $M$, Assembly Graph $\mathcal{G}_t=(\mathcal{V},\mathcal{E})$, Threshold $\lambda$}
		\State $S \leftarrow$ list()
		\If{$\left( \forall v \in \mathcal{V}: \  v.\textit{assembled-flag} == 1 \right)$}
		\State \textbf{return} $S$ \Comment{Exit: all parts assembled}
		\EndIf
		\State $\mathbf{y}\leftarrow M(\mathcal{G}_t)$
		\For{$i \leftarrow 1$ to $|\mathcal{V}|$}
		\If{$\mathbf{y}[i] < \lambda$}
		\State \textbf{continue}
		\EndIf
		\State $\mathcal{G}_{t+1} \leftarrow$ copy($\mathcal{G}_t$)
		\State $[\mathcal{V}_{t+1}]_i.\textit{assembled-flag} \leftarrow 1$ \Comment{assembled node $i$ }
		\State $S_{\ast} \leftarrow$ \textsc{Traverse-Tree($M, \mathcal{G}_{t+1}, \lambda$)}
		\For{$s$ in $S_{\ast}$}
		\State $s_{\ast} \leftarrow [i] + s$ \Comment{Add current part to the sequence}
		\State $S$.append($s_{\ast}$)
		\EndFor
		\EndFor
		\State \textbf{return} $S$
		\EndFunction
	\end{algorithmic}
\end{algorithm}

\subsubsection{Feasibility Prediction}
To address the issue from infeasible assemblies, we develop two schemes to infer the feasibility (defined in ~\ref{sec:problem_formulation}) of a given assembly: 
(1) We use the number of predicted complete sequences (output by Algo.~\ref{alg:03_walk_tree}) as an indicator for the feasibility of a given assembly. 
If no sequences were retrieved, the assembly is predicted as infeasible.
(2) We aggregate the features of all part nodes from a pre-trained GRACE with a \textit{mean-pooling} operation, creating a feature vector for the entire assembly graph. 
This feature vector is then used to train a binary classifier for feasibility prediction, where 
we analyze several classifiers i.e. Support Vector Machines (SVMs), Multi-layer Perceptrons (MLPs) and Nearest Neighbor.

\section{Experiment}
In this section, we first describe the experimental setup such as our dataset, evaluation metrics and implementation details.
To note that we use the term \textit{size} to describe the number of parts of an assembly without prior notice. 
We evaluate the Sequence Prediction under two experimental protocols with 4-fold cross-validation: {(1) \textbf{intra-sized}:} the assemblies in training and test set share \textit{the same} sizes; {(2) \textbf{inter-sized}:} the assemblies in training and test set have \textit{different} sizes, where there are two sub-protocols: Many-to-one and One-to-Many (detailed in \ref{sec:inter-sized})
The results on Feasibility Prediction are presented before the failure analysis and ablation study. 

\subsection{Experimental Setup}
\subsubsection{Dataset}
\begin{figure}[bt]
	\centering
	\includegraphics[width=1\linewidth]{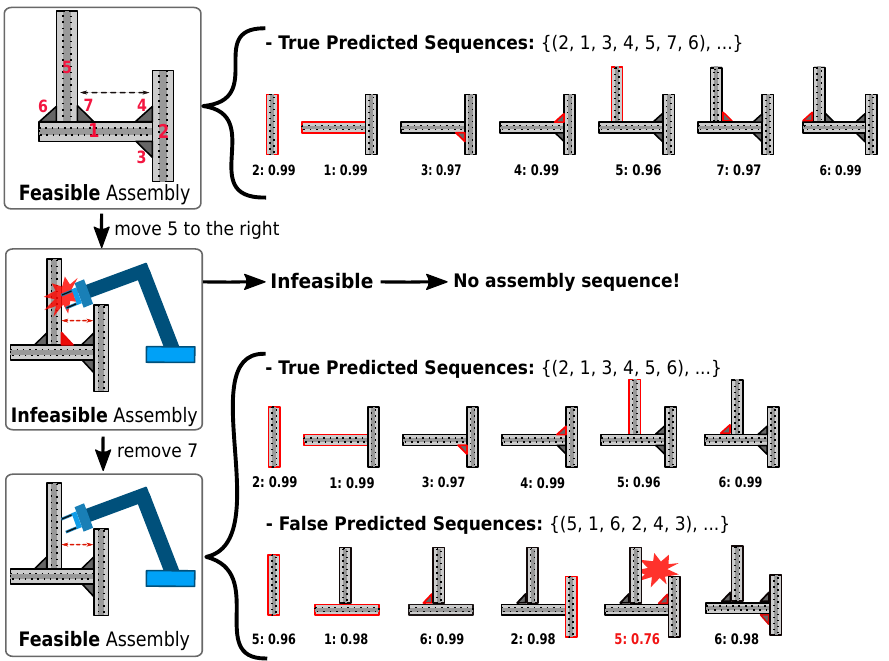}
	\caption{\textbf{Examples of \gls{asp} for Aluminum Assemblies.}
		We demonstrate the complexity of our task through the predictions for three different assemblies: starting from the top, there is a feasible assembly sequence predicted based on the assembly on the left. By decreasing the distance between part 5 and 4, it becomes infeasible due to limited space for the robot arm. Further, by removing part 7, with certain sequences such as the one shown in the figure, it is feasible. But this does not work with another, i.e. the false predicted sequence, because this one would cause collision.}
	\label{fig:asp_exp_aluminum_assemblies}
\end{figure}

We applied our in-house simulation software MediView to randomly generate data of synthetic aluminium assemblies whose sizes range from 3 to 7 (denoted by $A_i$, where $i$ is the size).
The simulation software was tasked with putting together the structures by brute-forcing all part orders, while considering the restrictions of part geometries or those imposed by the capabilities of a dual-armed robotic system \emph{KUKA LBR Med} (Fig.~\ref{fig:teaser}). More restrictions could be added to this environment in a future work, e.g. taking into account grasp planning.
An illustration of this process is given in Fig.~\ref{fig:asp_exp_aluminum_assemblies}.
The resulting data consists of the following amount per size: $A_3:5717$, $A_4:2464$, $A_5:6036$, $A_6:2865$, $A_7:431$. 

We post-processed the simulation output to obtain the \textit{Placement Action} (required during training) -- the next possible placement actions given a state of an assembly in a feasible sequence. 
In addition, we derive the \textit{Feasibility} of each assembly based on the number of ground truth sequences e.g. 0 indicates an infeasible assembly.

\subsubsection{Metrics}
We use the following metrics for the sequence prediction task:
(1)~\textbf{Step-by-Step AUC} examines our method's predictive performance to infer the parts that should be assembled next given the current state by comparing the ground truth binary labels with the predicted step scores.
For this purpose we use the common Precision-Recall curve w.r.t. $\lambda$ and finally deriving an \gls{auc} score.
(2)~\textbf{Complete-Sequence AUC} evaluates the ability to infer the entire set of ground truth sequences, since a step-by-step evaluation only partially displays our application\footnote{Consider a method that predicts the first 97 steps correctly and fails in the 98-th step for a 100-parts-assembly.}.
We use Information Retrieval (IR) Precision-Recall~\cite{croft2010search}, devised for set prediction evaluation, computed as ${\text{IR-Precision} = \lvert RET \cap REL \rvert / \lvert RET \rvert}$, ${\text{IR-Recall} = \lvert RET \cap REL \rvert / \lvert REL \rvert}$, 
where $RET$ are the retrieved sequences and $REL$ are the relevant sequences (i.e.\ ones in the ground truth set).
Here, again, we plot an IR Precision-Recall curve and derive an \gls{auc} score.
(3)~\textbf{\gls{pk}:} since in practice we only consider the highest scored predicted sequences, we also compute $\text{IR-Precision}$ while taking into account 
only the top-$k$ ones~\cite{herlocker2004evaluating}. 

For feasibility prediction, we use common binary classification metrics \emph{False Positive Rate}~$FPR$ and \emph{True Positive Rate}~$TPR$. In this setting, a positive instance is a feasible assembly and a negative an infeasible one.

\subsubsection{Implementation Details}\label{implementation}
We use PyTorch Geometric (PyG)~\cite{Fey_Fast_Graph_Representation_2019} to build the model which consists of $3$ surface blocks and $1$ part block with a latent-dimensionality of $94$.
For training, we choose a batch size of $256$ and a learning rate of $0.0022$ with Adam optimizer based on validation performance on $15\%$ of the training samples during hyper-parameter search.
Besides, we set the regularization weight in Eq.~\ref{eq:03_loss} to $0.3$ and length of positional encoding for node features to 16.
For the training and evaluation of the feasibility classifiers a balanced dataset is used. 
Our model includes $51.7 K$ trainable parameters and requires $4.06 \pm 0.15$ ms to infer the next feasible sequence step\footnote{Measured on NVIDIA GeForce GTX 1080.}.
More details are referred to our open-sourced code.

\subsection{Results} 
\subsubsection{Sequence Prediction for Intra-sized Assemblies}
The results are shown separately per assembly size (Tab.~\ref{exp:tab_knowledge_transfer_combined}), including the step-by-step and complete-sequence \gls{auc}, \gls{pk} scores for $k \in \{ 1, 2, 3 \}$ with a threshold of $0.5$. 
GRACE is able to perform perfectly on step prediction for all sizes. 
More relevant to our goal and more challenging than step prediction, our method can reach $1.0$ for small sizes (e.g. $3$ and $4$ parts) on the task of complete sequence prediction.
However, we can observe a slight drop for larger sizes (e.g. $5$, $6$ and $7$ parts), implying the greater complexity for large assemblies.
Hence, GRACE can effectively learn an useful inductive bias from our proposed graph representation when trained with similar sizes.
To note that, this performance has already reached that of the approach in~\cite{rodriguez2020pattern} and GRACE is able to generalize to larger sizes which the previous one is incapable of (see Tab. \ref{tab:comparison} for a qualitative comparison).

\begin{table*}[t]
	\centering
 	\caption{\textbf{Sequence Prediction Results for intra-sized and inter-sized (many-to-one) assemblies.}
	}
	\resizebox{\textwidth}{!}{
	\begin{tabular}{l | c c c c c | c c c c c}
		\toprule
		\multirow{1}{*}{Metrics} & \multicolumn{5}{c|}{Intra-sized} & \multicolumn{5}{c}{Inter-sized} \\
		 & $A_3$ & $A_4$ & $A_5$ & $A_6$ & $A_7$ & $A_3$ & $A_4$ & $A_5$ & $A_6$ & $A_7$ \\
		\midrule
		Step-by-Step AUC ($\uparrow$) & $1.00 \pm 0.00$ & $1.00 \pm 0.00$ & $1.00 \pm 0.00$ & $1.00 \pm 0.00$ & $1.00 \pm 0.00$  & $0.99 \pm 0.02$ & $1.00 \pm 0.00$ & $0.98 \pm 0.10$ & $0.98 \pm 0.00$ & $1.00 \pm 0.00$  \\
		\cmidrule(lr){1-11}
		Complete Sequence AUC ($\uparrow$) & $1.00 \pm 0.00$ & $1.00 \pm 0.00$ & $0.96 \pm 0.02$ & $0.93 \pm 0.03$ & $0.97 \pm 0.02$ & $0.97 \pm 0.03$ & $1.00 \pm 0.10$ & $0.87 \pm 0.03$ & $0.90 \pm 0.04$ & $0.95 \pm 0.07$ \\
		\cmidrule(lr){1-11}
		$P@1$ ($\uparrow$)& $1.00 \pm 0.00$ & $1.00 \pm 0.00$ &  $0.95 \pm 0.04$ & $0.96 \pm 0.06$ & $0.99 \pm 0.01$ & $0.99 \pm 0.01$ & $0.98 \pm 0.03$ & $0.90 \pm 0.09$ & $0.88 \pm 0.07$ & $0.96 \pm 0.05$ \\
		$P@2$ ($\uparrow$)& $1.00 \pm 0.00$ & $1.00 \pm 0.00$ &  $0.94 \pm 0.04$ & $0.95 \pm 0.07$ & $0.99 \pm 0.01$ & $0.99 \pm 0.01$ & $0.97 \pm 0.03$ &  $0.87 \pm 0.12$ & $0.87 \pm 0.07$ & $0.96 \pm 0.04$ \\
		$P@3$ ($\uparrow$)& -- & $1.00 \pm 0.00$ & $0.99 \pm 0.01$ & $0.95 \pm 0.08$ & $0.99 \pm 0.02$ & -- & $0.95 \pm 0.06$ & $0.93 \pm 0.10$ & $0.85 \pm 0.08$ & $0.96 \pm 0.04$ \\
		\bottomrule
	\end{tabular}}
	\label{exp:tab_knowledge_transfer_combined}
\end{table*}

\subsubsection{Sequence Prediction for Inter-sized Assemblies} \label{sec:inter-sized}
To comprehensively evaluate the \textit{generalization} ability of GRACE across different sizes, a distinct limitation of previous works~\cite{rodriguez2020pattern,wells2019learning},
we further design two more \textit{challenging} sub-protocols under the inter-sized protocol.
\begin{itemize}
	\item \textbf{Many-to-one}: GRACE is trained on assemblies of mixed sizes but $i$, i.e. $A_{\forall j \neq i}$, and tested on $A_{i}$.
	\item \textbf{One-to-many}: GRACE is trained on a single-sized dataset $A_{i}$ and tested on all the other, i.e. $A_{\forall j \neq i}$.
\end{itemize}

1. \textit{Many-to-one}: This setting is similar to the intra-sized one except that we excluded assemblies of the size evaluated at test time from the training set.
When comparing the results in this setting to the intra-sized ones (Tab.~\ref{exp:tab_knowledge_transfer_combined}), we observe a slight performance decrease in \gls{auc} on step and sequence prediction.
However, note that $P@1$ and $P@2$ can still reach $\sim 1.0$ for small sizes ($3$ and $4$ parts) and $\sim 0.9$ for large sizes, indicating that GRACE is capable on generalizing to assembly variants with different sizes that have not been seen before.

2. \textit{One-to-many}: This setting is an inverse version of the previous one, which is more challenging, since the amount and diversity of the training set are much lower than before\footnote{We do not perform this experiment on $A_7$, as there are relatively small amount of assemblies with $7$ parts in the dataset.}. 
The results (lower triangular block in Tab.~\ref{exp:tab_knowledge_transfer_inter_sized_one2many}) provide a clear pattern that GRACE is able to obtain comparably better results for assemblies with less parts.  
For instance, trained with only $A_5$, GRACE preforms well for $A_3$ and $A_4$ which is reasonable as the constraints guiding smaller assembly structures are \emph{contained} in larger ones.
This shows the \textit{generalization} capability and \textit{sample efficient} learning ability (trained on single size and worked on smaller sizes) of our method. 
Nevertheless, the performance drops for larger assemblies (see the upper triangular block in Tab.~\ref{exp:tab_knowledge_transfer_inter_sized_one2many}).
We hypothesize that an increasing amount of items introduces new constraints that are not covered by the training data.
Thus, there might be a critical number of items containing all possible constraints that if included in the training data can lead to an overall generalizing model.

\begin{table*}[t]
	\centering
 	\caption{\textbf{Sequence Prediction Results for inter-sized assemblies in one-to-many setting.}
	}
	\resizebox{\textwidth}{!}{%
		\begin{tabular}{l | c c c c c | c c c c c}
			\toprule
			\multirow{2}{*}{Training Set} & \multicolumn{5}{c|}{Step-by-Step AUC ($\uparrow$) on assemblies of various sizes} & \multicolumn{5}{c}{Complete Sequence AUC ($\uparrow$) on assemblies of various sizes} \\
			& $A_3$ & $A_4$ & $A_5$ & $A_6$ & $A_7$ & $A_3$ & $A_4$ & $A_5$ & $A_6$ & $A_7$ \\
			\midrule
			$A_4$ & $0.92 \pm 0.11$ & -- & $0.48 \pm 0.12$ & $0.41 \pm 0.11$ & $0.43 \pm 0.12$ & $0.93 \pm 0.09$ & -- & $0.28 \pm 0.12$ & $0.25 \pm 0.15$ & $0.25 \pm 0.15$ \\
			$A_5$ & $0.93 \pm 0.06$ & $0.89 \pm 0.07$ & -- & $0.78 \pm 0.14$ & $0.59 \pm 0.16$ & $0.83 \pm 0.07$ & $0.70 \pm 0.09$ & -- & $0.36 \pm 0.12$ & $0.24 \pm 0.07$ \\
			$A_6$ & $0.90 \pm 0.10$ & $0.89 \pm 0.11$ &  $0.93 \pm 0.04$ & -- & $0.59 \pm 0.16$ & $0.73 \pm 0.16$ & $0.68 \pm 0.24$ & $0.71 \pm 0.13$ & -- & $0.24 \pm 0.07$ \\
			\bottomrule
	\end{tabular}}
	\label{exp:tab_knowledge_transfer_inter_sized_one2many}
\end{table*}

\subsubsection{Feasibility Prediction}
\begin{figure}[tb]
    \centering
    \includegraphics[width=0.9\linewidth]{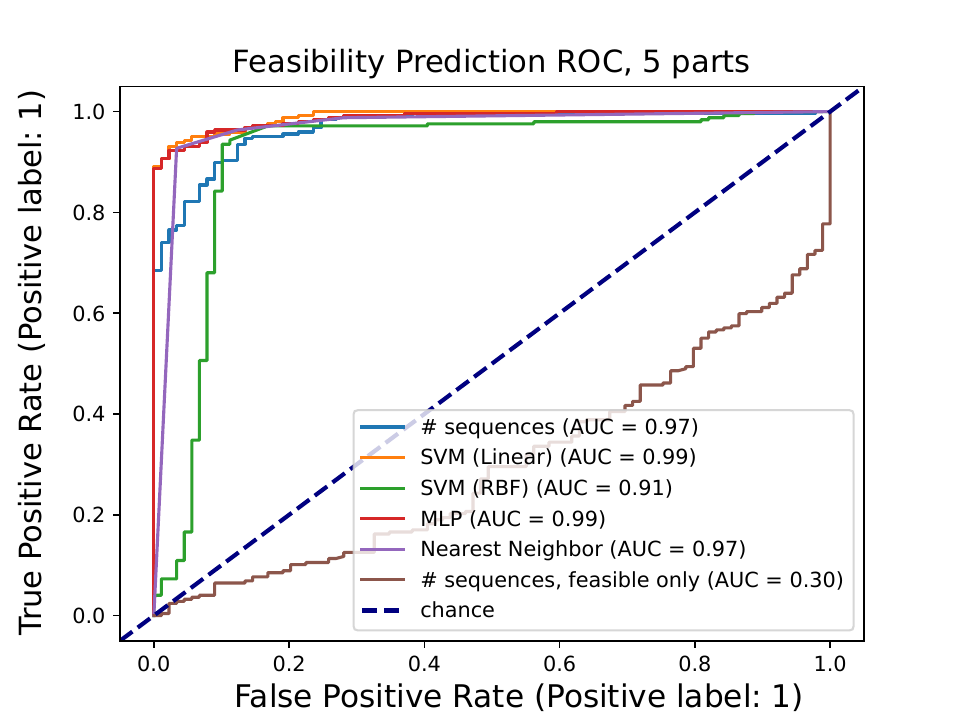}
    \caption{\textbf{Results of Feasibility Prediction.} Comparison of different proposed schemes for feasibility classification and an ablation study in which infeasible ones are unavailable based on assemblies with 5 parts.}
    \label{fig:04_classifier_comparision}
\end{figure}

In this setting, we examine the ability of our approach to detect infeasible assemblies.
For this experiment we consider GRACE trained on multiple sizes and test on the $A_5$ set.
As mentioned in~\ref{GRACE}, we compare the implicit approach via the number of predicted sequences (Algo.~\ref{alg:03_walk_tree}) and alternative schemes exploiting the graph representation of the pre-trained GRACE.
Therefore, we explore several binary classifiers i.e. SVMs, a Multi-layer Perceptron (MLP) and Nearest Neighbor.
As seen in Fig.~\ref{fig:04_classifier_comparision}, GRACE (\textit{\#sequences}) is able to detect infeasible assemblies (\gls{auc} of $0.97$).
However, training our method exclusively with feasible assemblies (\textit{\#sequences, feasilbe only}) results in a poor detection performance.
We hypothesize that by missing infeasible structures during training the method learns to always assemble an item leading to overconfidence.
Exploiting an additional scheme by adding one of the classifiers (except SVM with RBF kernel) maintains or even slightly improves the performance.

\subsection{Failure Analysis}\label{seq:limitation_analysis}
\begin{figure}[tb]
    \centering
        \includegraphics[width=1\linewidth]{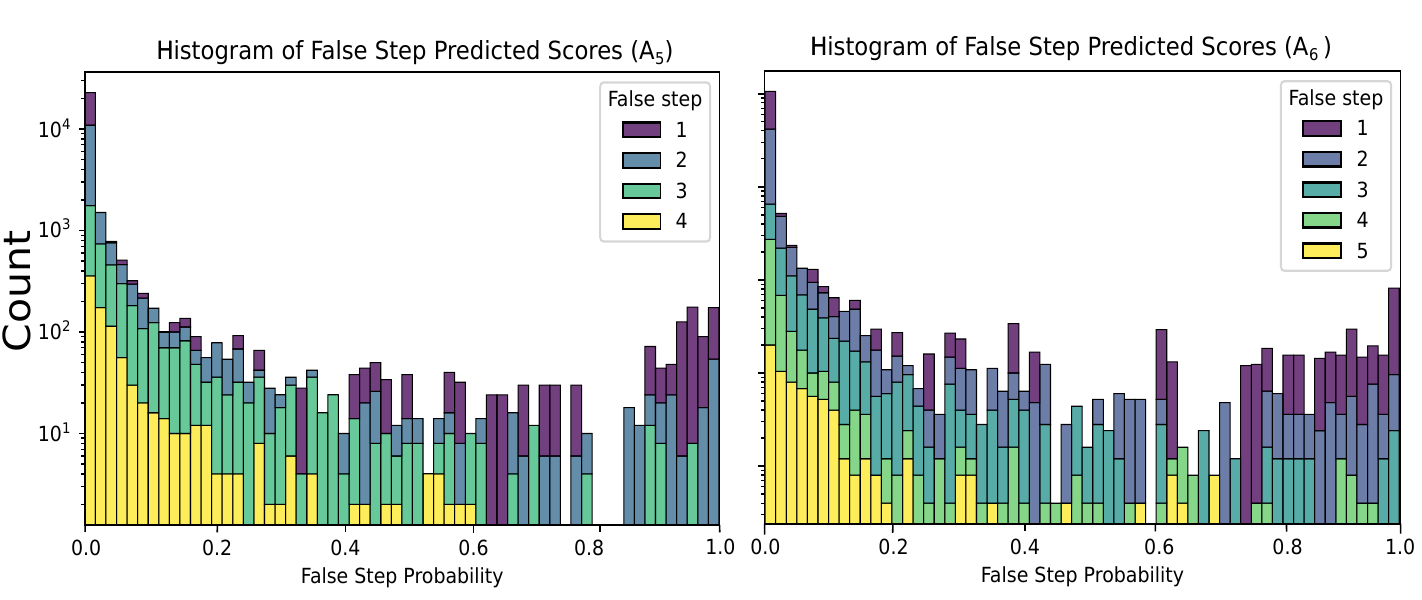}
    \caption{\textbf{Failure Analysis: false predicted sequences.} Histogram of predicted probability (for $A_5$ and $A_6$) in false steps reveals a drift in which GRACE is overconfident in mistakes preformed early. This is an evidence for an inherit bias in our training setting.}
    \label{fig:failure_analysis}
\end{figure}

To better understand the limitations of our method, we conduct an analysis of falsely predicted assembly sequences by our baseline model for $A_5$ and $A_6$.
Each of these false predicted sequences includes a \textit{false step}, i.e.\ action from which the sequence deviates from the corresponding ground truth sequences. 
Fig.~\ref{fig:failure_analysis} depicts the histogram of false steps binned by their predicted probability.
One can observe a large amount false steps performed in the beginning of the sequence (steps $1$ and $2$) with a high confidence.
On the other hand, wrong step predictions at the end of an assembly (steps $4$ and $5$) exhibit lower confidence scores. 
We hypothesize that this bias is a result of an inherit imbalance in our training setting. 
Our dataset samples could be thought of as nodes in a state tree, where earlier steps share state nodes closer to the root and later ones have independent nodes towards the leaves.
As each of these nodes is represented only once, there are fewer samples in the dataset attributed to earlier steps. 
This problem could be solved by balancing the training set based on the sequence step.

\subsection{Ablation Study}\label{seq:04_ablation}
In Assembly Graph (\ref{method:aseembly_graphs}), both the part and surface node embeddings contain a $16d$ sinusoidal positional encoding~\cite{vaswani2017attention}. 
We conducted an ablation study to investigate the impacts from the values and permutation order thereof based on $A_5$.
(1) \textbf{Values}:
Initializing the positional encoding with random values dramatically decrease the performance of our method (Tab.~\ref{tab:pos_encoding}).
We hypothesize that these positions introduce \textit{geometrical bias}, which is helpful in our task.
\begin{table}[t]
	\centering
 	\caption{Ablation study into the contribution of positional encodings to our method.}
	\begin{tabular}{l | c }
		\toprule
		Poisitional Encoding & $A_5$ AUC ($\uparrow$) \\
		\midrule
		Baseline, sinusoidal encoding~\cite{vaswani2017attention} & 0.94 \\
		Random values & 0.37 \\
		No encodings & 0.07 \\
		\midrule
		Part permutations (test time) & 0.60 \\
		Surface permutations (test time) & 0.27 \\
		Part permutations (training and test time) & 0.97 \\
		Surface permutations (training and test time) & 0.10 \\
		\bottomrule
	\end{tabular}
        \label{tab:pos_encoding}
\end{table}
(2) \textbf{Permutation Order}:
We number assembly parts and surfaces in a constant order. 
Parts are counted beginning from the one closest to the environment origin. 
Surfaces, on the other hand, are always numbered clockwise, starting from the respective part top. 
Permuting both part and surface orders \textit{only} at test time causes severe performance degradation, indicating constant numbering during training harms the model's ability to generalize (Tab. ~\ref{tab:pos_encoding}).
Interestingly, allowing permutations for only part order can boost the performance while this is not the case for surface permutations.
This demonstrates the importance of these features for the network to extract information from the parts' geometrical structure.

\section{Conclusion}
In this work, we addressed the RASP problem with a learning-based framework. 
Concretely, we propose a graph representation, called Assembly Graphs for the aluminum profile assemblies, which is flexible to represent different 2d structures and meanwhile agnostic to rotation and mirroring.
Based on this, a novel policy network -- GRACE is introduced to extract meaningful information for assembly sequence prediction. 
Extensive experiments in simulation verify the capability of transferring knowledge between different assembly tasks, on which previous methods fall short. 
Further, our method can generalize knowledge gained on larger assemblies and then apply it to smaller ones. 
Last but not least, it is worth to mention, though only validated in simulation, our method should address the challenges during the real-world deployment like not finding a valid motion or a feasible grasping point if these cases are enclosed in the training data and learned to reject by GRACE.
Meanwhile encouraged by the superior results on objects with simple geometries, our holistic graphical method lays a solid basis for handling complex 3d objects like curve blocks in the future. 

\section*{Acknowledgments}
The paper received partial funding by the DLR internal project Factory of the Future Extended (FoF-X).
Jianxiang Feng is supported by the Munich School for Data Science (MUDS). Rudolph Triebel is a member of MUDS.





\bibliographystyle{IEEEtran}
\bibliography{IEEEabrv,references}

\end{document}